\documentclass{article}
\usepackage{ijcai25}

\usepackage{times}
\usepackage{soul}
\usepackage{url}
\usepackage[hidelinks]{hyperref}
\usepackage[utf8]{inputenc}
\usepackage[small]{caption}
\usepackage{graphicx}
\usepackage{amsmath}
\usepackage{amsthm}
\usepackage{booktabs}
\usepackage{algorithm}
\usepackage{algorithmic}
\usepackage[switch]{lineno}

\title{Exploring Semantic Masked Autoencoder for Self-supervised \\ Point Cloud Understanding}

\author{
    Yixin Zha*, Chuxin Wang*, Wenfei Yang, Tianzhu Zhang
    \affiliations
    University of Science and Technology of China / Deep Space Exploration Lab
    \emails
    \{zyxcn, wcx0602\}@mail.ustc.edu.cn \thanks{Equal Contribution}
}

\usepackage{newfloat}
\usepackage{listings}
\usepackage{booktabs}
\usepackage{multirow}
\usepackage{pifont}
\usepackage{xcolor}
\usepackage{colortbl}

\begin{document}

\maketitle
\begin{abstract}

% Point cloud understanding aims to acquire robust and general feature representations from unlabeled data. 
% Masked point modeling-based methods have recently shown significant performance across various downstream tasks. 
% These pre-training methods rely on random masking strategies to establish the perception of point clouds by 
% restoring corrupted point cloud inputs. 
% However, due to the masking strategy, these methods fail to capture reasonable component semantics by the self-supervised models. 
% To address this issue, we propose a Semantic Masked Autoencoders, 
% which comprises two main components, a prototype-based component semantic modeling module and a component semantic-enhanced masking strategy. 
% Specifically, in the component semantic modeling module, 
% we design a component semantic guidance mechanism to direct a set of learnable prototypes 
% in capturing the semantics of different components from objects. Leveraging these prototypes, 
% we develop a component semantic-enhanced masking strategy that addresses the limitations of random masking 
% in effectively covering complete component structures. Furthermore, we introduce a component semantic-enhanced prompt-tuning strategy, 
% which further leverages component semantic modeling prototypes to improve the performance of the pre-trained model in downstream tasks. 
% Extensive experiments conducted on datasets such as ScanObjectNN, ModelNet40, and ShapeNetPart demonstrate the effectiveness of our proposed modules.

Point cloud understanding aims to acquire robust and general feature representations from unlabeled data. Masked point modeling-based methods have recently shown significant performance across various downstream tasks. These pre-training methods rely on random masking strategies to establish the perception of point clouds by restoring corrupted point cloud inputs, which leads to the failure of capturing reasonable semantic relationships by the self-supervised models. To address this issue, we propose Semantic Masked Autoencoder, which comprises two main components: a prototype-based component semantic modeling module and a component semantic-enhanced masking strategy. Specifically, in the component semantic modeling module, we design a component semantic guidance mechanism to direct a set of learnable prototypes in capturing the semantics of different components from objects. Leveraging these prototypes, we develop a component semantic-enhanced masking strategy that addresses the limitations of random masking in effectively covering complete component structures. Furthermore, we introduce a component semantic-enhanced prompt-tuning strategy, which further leverages these prototypes to improve the performance of pre-trained models in downstream tasks. Extensive experiments conducted on datasets such as ScanObjectNN, ModelNet40, and ShapeNetPart demonstrate the effectiveness of our proposed modules.

\end{abstract}

% Introduction
\section{Introduction}

As a direct and accurate representation of real-world objects and environments, point clouds attract extensive attention from the academic community. 
In recent years, generous fully-supervised point cloud representation methods~\cite{qi2017pointnet,qi2017pointnet++,zhao2021point,wang2019dynamic,zhao2021point} 
have been proposed and have shown promising performance in various 3D shape analysis and scene understanding tasks. 
However, these methods often rely on large annotated point cloud datasets, which is time-consuming and requires a significant amount of manpower and economic cost.

\begin{figure}[t] % h 选项表示将图片放在当前位置，而不是浮动到页面的顶部或底部
 \centering % 居中显示图片
\hspace{-0.4cm}
\includegraphics[width=0.45\textwidth]{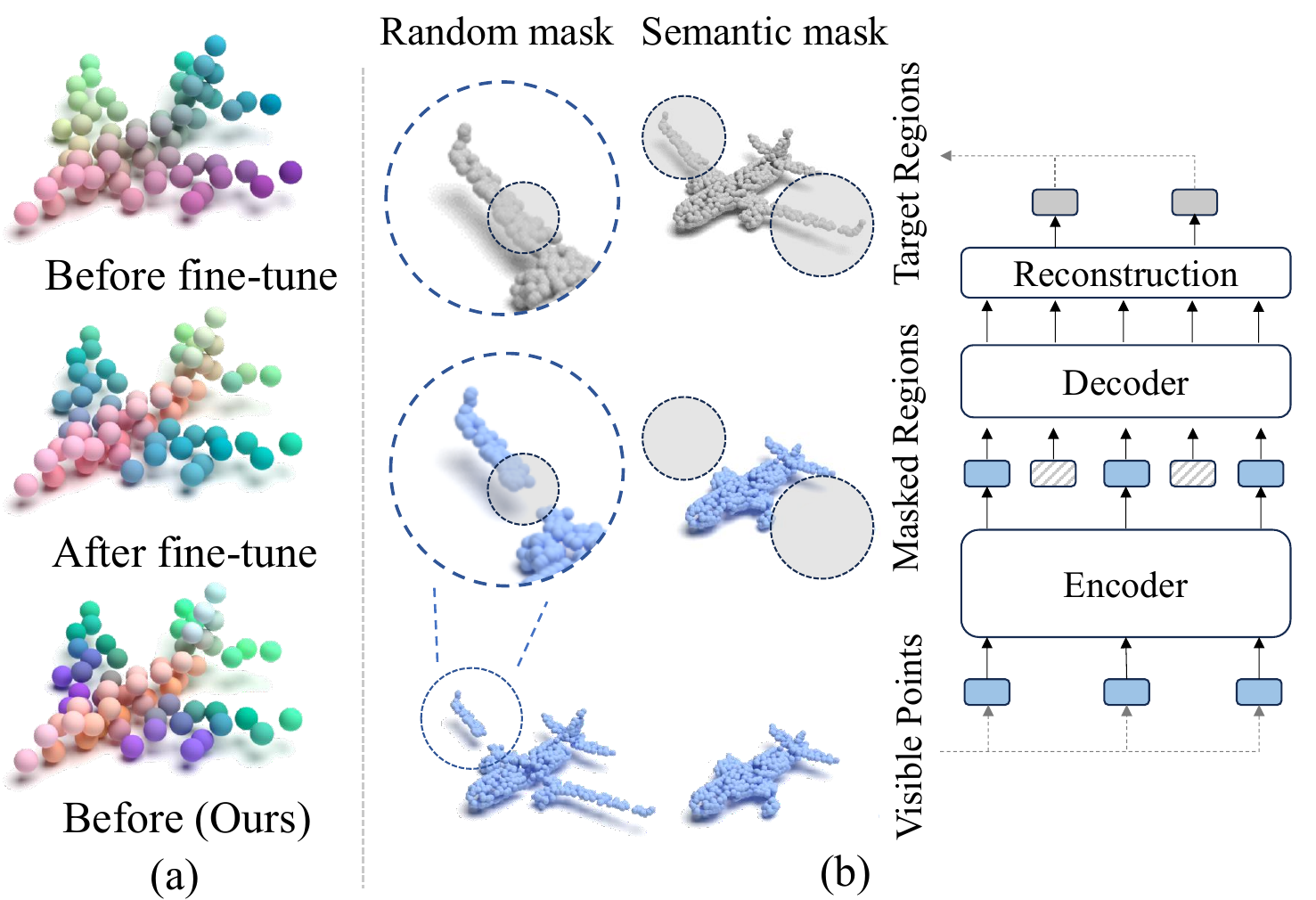} % 插入图片，并设置宽度为页面宽度的一半
% \vspace{-3mm}
\caption{(a) \textbf{Features distribution.} We visualize the features before and after fine-tune. The feature colors are transformed into feature space using PCA, where the same color indicates feature consistency.
 (b) \textbf{Different mask and corresponding reconstruction processes.} Random masking masks only local blocks, while semantic masking masks complete components based on semantics (wings). Points in gray circles are masked regions.
 }  % 添加图片标题
 \label{fig:0} % 添加图片标签，方便交叉引用
%  \vspace{-3mm}
\end{figure}

Inspired by self-supervised learning~\cite{radford2018improving,radford2019language,brown2020language} in image processing and natural language, numerous self-supervised 3D representation learning methods have been proposed. 
These methods can learn meaningful representations on unlabeled point cloud datasets and enhance model performance on downstream tasks through fine-tune. 
Existing self-supervised methods primarily follow two research paths, contrastive learning method~\cite{xie2020pointcontrast,zhang2021self,dong2022autoencoders} 
and generative methods~\cite{yu2022point,2022Masked,zhang2023learning}. 
Contrastive learning methods guide the model in learning discriminative features by distinguishing positive and negative samples. 
For example, PointContrast~\cite{xie2020pointcontrast} constrains the consistency between the same points in different views. 
CrossNet~\cite{wu2023self} conducts cross-modal contrastive learning between point clouds and their corresponding rendered images. 
Contrastive learning methods yield satisfactory results, but their performance heavily relies on carefully designed positive and negative sample pairs.
Motivated by BERT~\cite{devlin2018bert} and MAE~\cite{he2022masked}, 
generative methods mainly adopt the Masked Point Modeling (MPM) to encourage models to infer the randomly masked regions with the visible regions, 
thereby guiding the model to learn the relationships between point cloud patches in the process, such as PointMAE~\cite{2022Masked}, PointMamba~\cite{liang2024pointmamba} and PointBERT~\cite{yu2022point}
% PointMAE~\cite{2022Masked} employs the random mask strategy and reconstructs the structure of the masked regions. 
% PointMamba~\cite{liang2024pointmamba} replaces transformer with SSM-based model~\cite{gu2023mamba} to capture visible regions inter-components relationships with random masking.
% PointM2AE~\cite{zhang2022point} conducts hierarchical encoders to extract multi-scale point cloud features while still adopting a random masking strategy.

Despite the success of MPM methods, 
they all fail to capture component semantics,
which limits the generalization ability of the pre-trained model.
As shown in Figure~\ref{fig:0}(a), 
without fine-tune, 
the model's output features exhibit strong positional instead of semantic correlation. For example,
the wings features are different before fine-tune due to they are not adjacent in 3D space, even though their structures are similar. 
After analysis, we argue that the strong positional correlation of features is caused by the random masking strategy.
Random masking causes residual local structure, 
which encourages models to infer masked regions with only adjacent patches rather than component semantic relationships.
For instance, as shown in Figure~\ref{fig:0}(b), 
the wings of airplanes are large flat surfaces, if only a few blocks are masked, 
the network will reconstruct these blocks with residual local structural information (i.e., flat surface).
This nature of random masking weakens the complete component semantic relationship reasoning ability of pre-trained models. 

Motivated by the analysis above, 
we believe that complete components of objects should be masked based on semantic priors as shown in Figure~\ref{fig:0}(b),
this encourages models to infer masked complete components based on visible point clouds, 
which enables model to capture semantic relationships between visible and masked components through reconstruction. 
However, to achieve this, we need to address two technical challenges:
% \emph{\textbf{(1) Perform masking based on semantic priors.}}
% We should mask complete components of objects according to semantic priors,
% this enables the network to learn the semantic relationships between the visible and masked complete components through reconstruction.
\textbf{(1) Explicitly Model component semantics during pre-training.}
To acquire semantic priors for semantic masking, the model should be able to explicitly model the semantics of local components during the pre-training phase, 
which is quite challenging without any supervision. 
\textbf{(2) Leverage component semantics for pre-training.}
With the explicit modeling of component semantics, how to leverage component semantics to perform point cloud understanding tasks is a problem worth exploring.

To achieve above goals,
we propose Semantic Masked Autoencoder,
which comprises two main components: a Prototype-based Component Semantic Modeling (PCSM) module and a Component Semantic-enhanced Masking (CSeM) strategy.
Specifically, in the PCSM,
we use learnable prototypes to capture different component semantics with complete point cloud inputs,
and guide these prototypes to model reasonable local semantics by reconstructing input point clouds with themselves.
Leveraging these component semantic prototypes, we develop a CSeM strategy that addresses the limitation of random masking in effectively covering complete components. 
We partition the complete point clouds into several components with different local semantics, and instead of random masking on the whole point clouds, 
we first completely mask a few of these components to prevent local structure information leakage, 
and then perform random masking on each remaining component. 
Furthermore, we introduce a Component Semantic-enhanced Prompt-tuning (CSeP) strategy, 
which leverages component semantic prototypes to improve the performance of the pre-trained model during the fine-tuning stage. 

In summary, our main contributions are as follows: 
(1) We propose Semantic Masked Autoencoder to improve existing MPM methods effectiveness
by explicitly modeling component semantics during pre-training and performing masking based on semantic priors.
(2) We introduce a PCSM module to model local semantics without supervision.
We design a CSeM to improve pre-trained models's ability to model local semantics.   
Besides, we employ the semantic-enhanced prototypes to improve the pre-trained models performance in downstream tasks. 
(3) Extensive experiments on datasets such as ScanObjectNN, ModelNet40, and ShapeNetPart demonstrate the effectiveness of our proposed modules.

% %%%%%%%%%% RelatedWork

\section{Related Work}
\subsection{Contrastive-based Point Cloud Understanding}
Self-supervised Learning has achieved great success in many fields such as NLP and computer vision.
The objective of PCSSL is to extract robust and general features from unlabeled data and achieve superior performance in downstream tasks.
Recently, a substantial amount of methods on self-supervised representation learning for point clouds has been proposed, 
demonstrating effectiveness in various shape analysis and scene understanding tasks. 
Existing self-supervised 3D representation learning methods are primarily follow two research paths: 
contrastive learning methods~\cite{xie2020pointcontrast,zhang2021self,dong2022autoencoders} and generative methods~\cite{yu2022point,2022Masked,zhang2023learning}. 
Inspired by the contrastive approaches,
PointContrast~\cite{xie2020pointcontrast} first explores learning 3D representations by constrainting the consistency between the same points in different views.
CrossPoint~\cite{afham2022crosspoint} learns point cloud representations by contrast learning, and then performs further cross-modal contrast learning.
CrossNets~\cite{wu2023self} conducts cross-modal contrastive learning between point clouds and their corresponding rendered images. 
However, contrastive learning methods heavily rely on meticulously designed positive and negative samples, which limits their development. 

\subsection{MAE-based Point Cloud Pre-training}
Generation methods~\cite{vincent2008extracting,devlin2018bert,radford2018improving,zhang2022point}
typically rely on an autoencoder to learn the latent features of the data by reconstructing the original inputs. 
Masked autoencoders (MAE)~\cite{he2022masked} try to recover the origin inputs from corrupted inputs, which allows the model to learn more robust features.
Inspired by MAE, MAE-based point cloud pre-training methods have been widely proposed, and can be divided into two categories, 
single-modal~\cite{zhang2022point,2022Masked} and cross-modal~\cite{dong2022autoencoders,qi2023contrast,zhang2023learning,qi2023contrast} methods. 
Point-MAE~\cite{2022Masked} pioneers the use of masked autoencoders for point cloud understanding. 
It divides point clouds into patches and employs mini-PointNet to extract patch embeddings, 
then a random mask reconstruction is performed with standard transformers. 
PointM2AE~\cite{zhang2022point} proposes a multi-scale masking strategy, 
but still relies on a global random masking strategy. 
Subsequent methods mainly focus on using cross-modal knowledge to aid point cloud understanding.
For instance, ACT~\cite{dong2022autoencoders} utilizes a pre-trained ViT~\cite{dosovitskiy2020image} as a teacher network to guide the learning of the student network.
Recon~\cite{qi2023contrast} learns from both generative modeling teachers and cross-modal contrastive teachers through ensemble distillation. 
Other MAE-based methods~\cite{chen2023pimae,yang2023gd,tian2023geomae} focus on using scene and LiDAR point clouds for pre-training, specifically for detection tasks. 
All above methods are rely on global random masking strategy, 
however, the random masking fails to mask complete components, which weakens the effectiveness of pre-text tasks.
Researchers have noted the drawbacks of the random masking. I2P-MAE~\cite{zhang2023learning}uses a cross-modal approach, leveraging rich 2D information to select crucial local structures as visible parts.
However, the need for extensive image and text datasets and additional computational costs limit its practical application.

% %%%%%%%%%%%%% Method

\section{Methodology}
\begin{figure*}[ht] % h 选项表示将图片放在当前位置，而不是浮动到页面的顶部或底部
    \centering % 居中显示图片
    \includegraphics[width=1\textwidth]{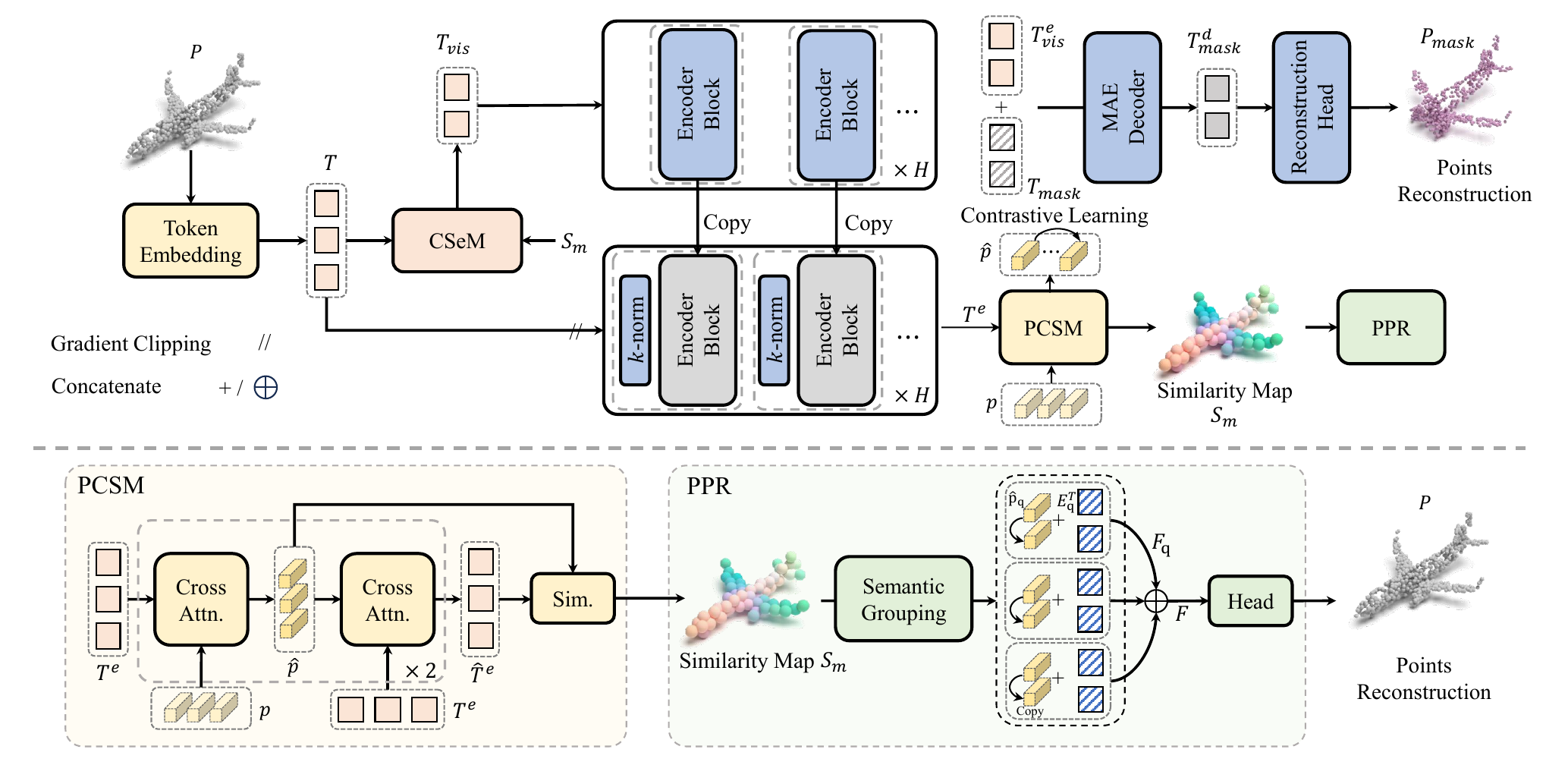} % 插入图片，并设置宽度为页面宽度的一半
    \caption{\textbf{Pipeline of our proposed framework.}
    The complete tokens $T$ are input into the encoder to acquire $T^e$, which are fed into PCSM to generate local semantic-enhanced prototypes $\hat{p}$ and enhanced tokens $\hat{T^e}$.
    The PPR guides the learning of $\hat{p}$ to acquire informative prototypes, and $S_m$ are solved before Semantic Grouping.
    With $S_m$, we can utilize CSeM to generate semantic-correlated masks. The generated masks are capable of covering complete point cloud components.
    The remaining tokens $T_{vis}$ are fed into the MAE-based architecture for self-supervised pre-training. 
    \textbf{The gray blocks indicate that the corresponding structures are frozen.}}  % 添加图片标题
    \label{fig:2} % 添加图片标签，方便交叉引用
    % \vspace{-3mm}
\end{figure*}

The overall pipeline of the proposed Semantic Masked Autoencoder is shown in Figure~\ref{fig:2}.
We first introduce the MAE-based 3D architecture for point cloud masked autoencoding. 
Then, we show the details of two main components, 
a prototype-based component semantic modeling module(PCSM) and a component semantic-enhanced masking strategy(CSeM).
And the component semantic-enhanced prompt-tuning(CSeP) is presented at the end of the methodology section.

\subsection{MPM-based 3D architecture}
Before introducing our proposed approaches, we will first outline the general architecture of MAE-based methods.
The 3D point cloud masked autoencoding consists of a token
embedding module, an encoder-decoder architecture, and a
head for reconstructing masked 3D points.
Our approaches can be implemented within any MAE-based method.

\textbf{Token Embedding.}
Given a raw point cloud $P \in \mathcal{R} ^{N \times 3}$,
Furthest Point Sampling (FPS) is applied to downsample the point number from $N$ to $G$, resulting in 
$P^T \in \mathcal{R} ^{G \times 3}$.
Then, $k$ Nearest-Neighbour ($k$-NN) is adopted to
search the $k$ neighbors for each downsampled point, and
their features are aggregated via a mini-PointNet \cite{qi2017pointnet} to obtain
$G$ point tokens. We formulate these tokens as $T \in \mathcal{R} ^{G \times C}$ , where $C$ denotes the feature dimension.
Simultaneously, the position embeddings of $P^T$ are generated by a linear layer, denoted as $E^T \in \mathcal{R}^{G \times C}$.

\textbf{Encoder-Decoder Architecture.}
To build the pre-text task targets, MAE-based methods usually mask the point tokens with a preset ratio,
only visible tokens can be fed into encoders. We formulate visible tokens as $T_{vis} \in \mathcal{R}^{G_{vis} \times C}$ ,
where $G_{vis}$ denotes the visible token number. 
The specific framework of the encoder could be different structures such as Transformer~\cite{2022Masked,yu2022point} or Mamba ~\cite{liang2024pointmamba}.
After encoding, the encoded $T^e_{vis}$ are concatenated with a set of shared learnable masked tokens $T_{mask} \in R^{M_{mask} \times C}$ , and then inputted
into a corresponding decoder, where $G_{mask}$ denotes the
masked token number and $G = G_{mask} + G_{vis}$. In the
decoder, the masked tokens learn to capture informative geometric cues from visible ones and reconstruct
the masked 3D regions.

\textbf{Points Reconstruction.}
With the decoded point tokens $\{ T^d_{vis}, T^d_{mask}\}$  and corresponding position embedding $\{E^T_{vis}, E^T_{mask}\}$
, $T^d_{mask}$ are utilized to reconstruct 3D regions of the masked tokens.
A reconstruction head usually consists of 
a single linear projection layer which is adopted to predict
$P_{mask} \in R^{G_{mask}\times k \times 3}$,
the ground-truth 3D points of
the pre-text task. The reconstruction loss \cite{fan2017point} can be formulated as:
\begin{equation}
    L_{3D} = \frac{1}{G_{mask}} \mathrm{Chamfer}(\mathrm{M_{3D}}(T^d_{mask}), P_{mask})
\end{equation}
where $\mathrm{M_{3D}(\cdot)}$ denotes the reconstruction head.

\subsection{Component Semantic Modeling}
We adopt learnable prototypes to adaptively capture semantics of different local structures.
The updated prototypes then support the subsequent CSeM approach.
We utilize the token embedding module and encoder from the MAE-based 3D architecture to obtain the complete point cloud features 
$T^e \in \mathcal{R}^{G \times C}$ and position embeddings $E^T$ of downsampled points $P^T$,
the learnable prototypes $p \in \mathcal{R}^{Q \times C}$ are updated with $T^e$ in cross-attention mechanism. 
To guide prototypes for semantic modeling of object components without additional modal information,
we reconstruct the entire point cloud with $p$ and $E^T$.

\textbf{Prototype-based Component Semantic Modeling.}
To adaptively model the point cloud components, we design a set of learnable prototypes to capture local semantics. 
As shown in Figure~\ref{fig:2}, given $T^e$,
we utilize a set of learnable prototypes $p$ to dynamically
aggregate these tokens. With the assistance of the
cross-attention mechanism, the aggregating process can be
described as follows:
\begin{equation}
    \hat{p} = \mathrm{Softmax}(\frac{p(T^e)^T}{\sqrt{d}})T^e
\end{equation}
where $\hat{p} \in \mathcal{R}^{Q \times C}$ represents the updated prototype features. 
After updating, we adopt these prototypes to enhance tokens $T^e$ in the another cross-attention layer,
enable each token to capture the semantic information of its corresponding component. The enhanced tokens can be formulated as $\hat{T}^e$.
During exploration, we find that the prototypes updated with origin encoders fail to capture the semantics of dispersed structures,
such as the wings of the airplane. 
To expand the encoder's receptive field, we introduce a non-parametric $k$-norm process inspired by Point-NN ~\cite{zhang2023parameter} to enhance the token's perception of distant local structures.
% With an encoder with $H$ stages, 
% we apply $k$-NN on $P^T$ combined with every stage's output $T^e_h$ to acquire token groups $K^e_h$,  
% then update each token $t^e_h$ with its corresponding token group $k^e_h$ following the local feature aggregation process.
The details of $k$-norm are presented in supplementary materials.

\textbf{Prototype-based Point Reconstruction(PPR).} 
To guide prototypes model reasonable local semantics of point clouds,
we first divide tokens into several groups, with each group being associated with a prototype that corresponds to a specific component.
Then, we reconstruct each component with its corresponding prototype.

To assign tokens according to their semantics and perform \textbf{Semantic Grouping}, we first figure out the scaled-product similarity between $\hat{T}^e$ and $\hat{p}$.
The similarity calculation can be described as follows:
\begin{equation}
    S_m = \mathrm{Softmax}(\frac{\hat{T}^e(\hat{p})^T}{\sqrt{d}})
    \label{eq:3}
\end{equation}
According to $S_m \in \mathcal{R}^{G \times Q}$, 
each token has a corresponding similarity score with all prototypes.
The highest similarity score reflects which component each token belongs to.
$E^T$ is the position embedding of tokens,
and we associate each position embedding with the corresponding prototype along with $S_m$.
% and the $E^T$ is partitioned to $Q$ groups $E^T_Q$. 
% We formulate the $q$-th position embedding groups as $E^T_q$, and each embedding in $q$-th groups as $e^T_q$.
To guide prototypes model reasonable local semantics, as shown in Figure~\ref{fig:2}, if the number of tokens in $q$-th group is $r$ ($E^T_q \in \mathcal{R}^{r \times C}$ ), 
we repeat $\hat{p}_q \in \mathcal{R}^{1 \times C}$ for $r$ times and then concatenate it with $E^T_q$ to forme $F_q \in \mathcal{R}^{r \times 2C}$.
All $F_q$ are concatenated to acquire $F \in \mathcal{R}^{G \times 2C}$, and we use a non-linear projection layer and a linear layer with $F$ to reconstruct raw points $P$.
The reconstruction loss can be formulated as:
\begin{equation}
    L_{proto} = \frac{1}{G} \mathrm{Chamfer}(\mathrm{M_{3D}}(F), P)
\end{equation}
where $\mathrm{M_{3D}(\cdot)}$ denotes the reconstruction head.
To guide different prototypes to focus on components with different semantics,
we utilize a contrastive loss on the updated prototypes $\hat{p}$,
as follows:
\begin{equation}
    \mathcal{L}_{cont} = -\sum_{i=1}^Q \mathrm{log}(\frac{\mathrm{exp}(d(\hat{p}_i, \hat{p}_i)/\epsilon)}{\sum_{j=1}^Q \mathrm{exp}(d(\hat{p}_i, \hat{p}_j)/\epsilon)})
\end{equation}
where $d(\cdot)$ is a distance measurement and $\epsilon$ is the temperature in contrastive learning. 

\begin{figure}[ht] % h 选项表示将图片放在当前位置，而不是浮动到页面的顶部或底部
    \centering % 居中显示图片
    \hspace{-0.1cm}
    \includegraphics[width=0.45\textwidth]{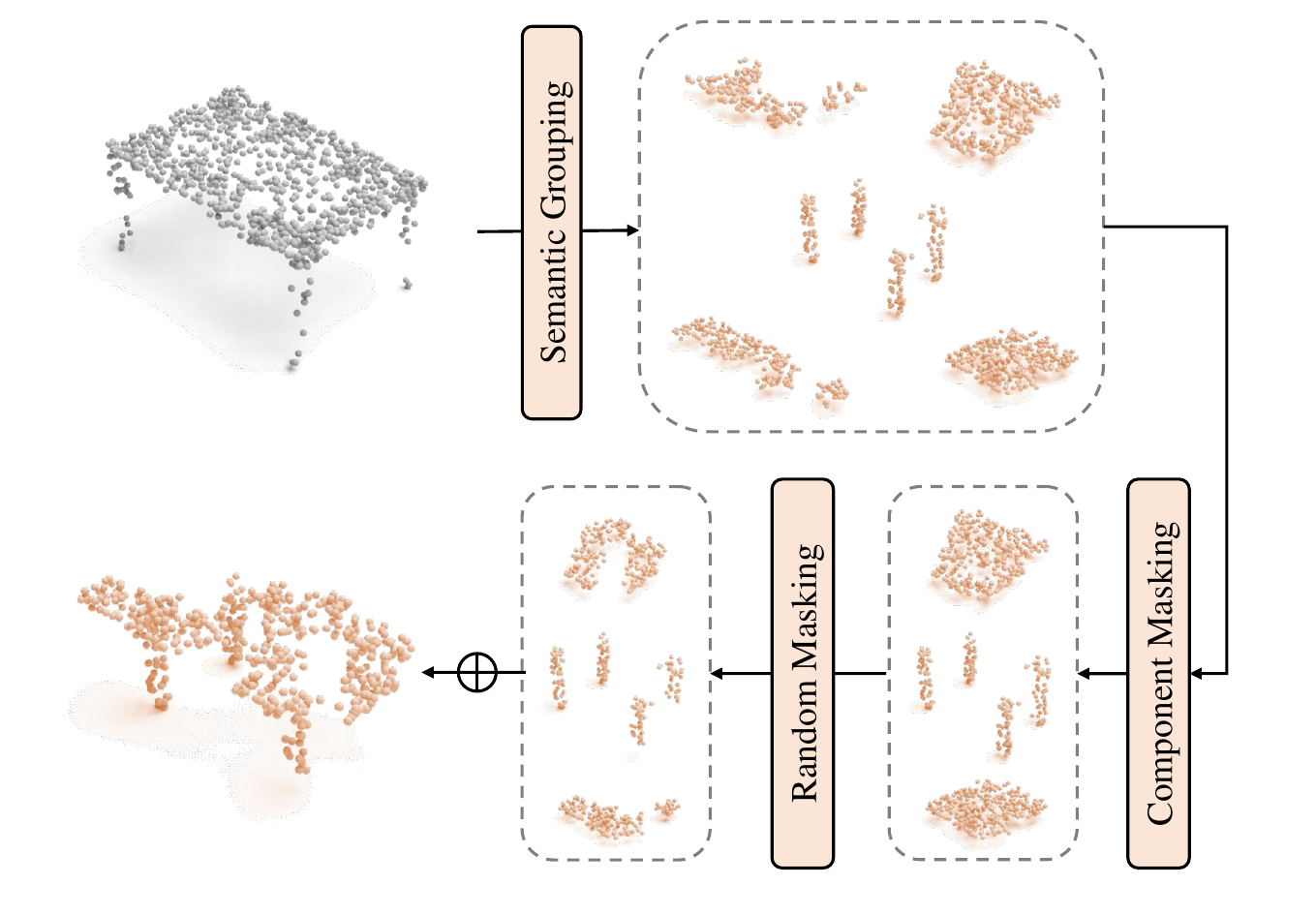} % 插入图片，并设置宽度为页面宽度的一半
    \caption{\textbf{Details of CSeM.} We first segment point clouds into several components according to local semantics and random select a few of them as masked components. Then we adopt random masking on each remaining component separately.
    }  % 添加图片标题
    \label{fig:3} % 添加图片标签，方便交叉引用
    % \vspace{-3mm}
\end{figure}

\begin{figure}[ht] % h 选项表示将图片放在当前位置，而不是浮动到页面的顶部或底部
    \centering % 居中显示图片
    % \vspace{-3mm}
    \includegraphics[width=0.45\textwidth]{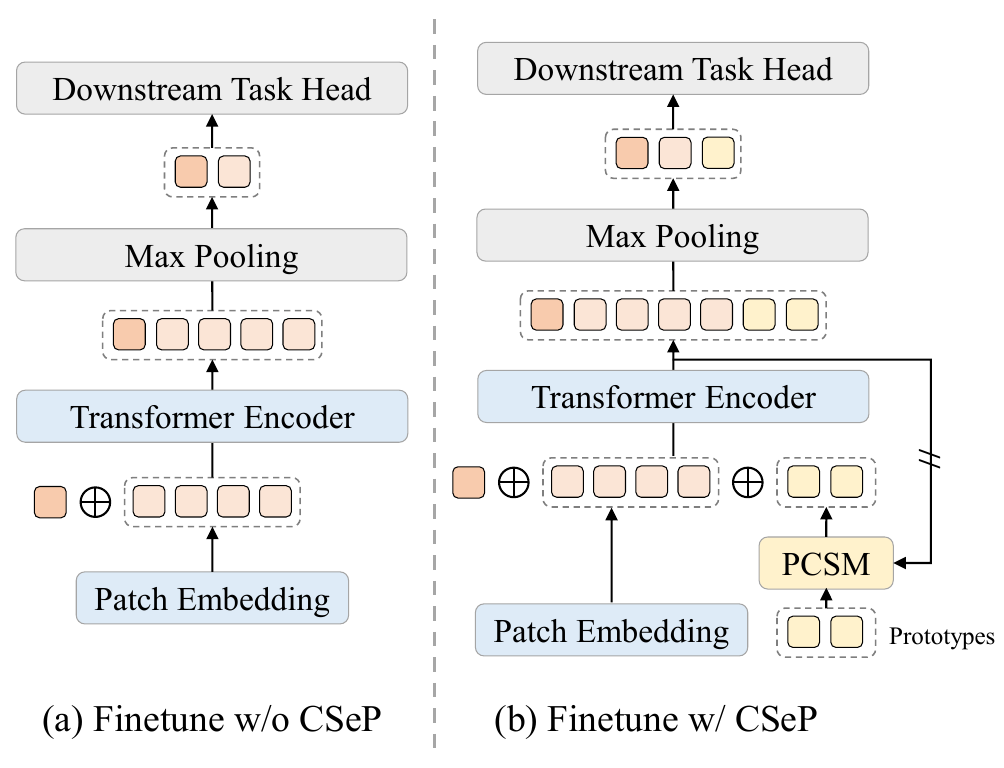} % 插入图片，并设置宽度为页面宽度的一半
    \caption{\textbf{Details of CSeP.} (a) The figure illustrates a structure widely used for downstream classification tasks. 
    (b) In the figure, we introduce the PCSM module during the finetuning stage to generate semantic-enhanced prototypes.
    }  % 添加图片标题
    \label{fig:4} % 添加图片标签，方便交叉引用
    % \vspace{-3mm}
\end{figure}

\subsection{Component Semantic-enhanced Masking Strategy}
% $T^e_Q \in \mathcal{R}^{r \times C}$
In MAE-based 3D architecture, point clouds are divided into several patches by FPS and $k$-NN.
Then, according to the token embedding module in MAE-base 3D architecture, tokens $T$ is generated via mini-Pointnet.

To perform semantically correlated mask, we first divide these tokens into different groups according to their semantics.
Specifically, with $S_m$ obtained from Formula~\ref{eq:3}, we follow the same grouping strategy to acquire semantical token groups.
Each token $t$ belongs to a specific component, which corresponds to the prototype with highest similarity score to $t$. 
The point cloud tokens $T^e$  are segmented into $Q$ groups along with $S_m$.
To mask the complete local structures, we randomly select some of these token groups $T^e_q \in \mathcal{R}^{r \times C}$ as masked components.
Besides, according to our experiments,  
directly masking components based on semantics fails to obtain optimal results. 
As shown in Figure~\ref{fig:5}, the components are extensive point groups, we believe that the massive masking creates a significant information gap between the model's inputs and ground truth, 
making it difficult for the model to learn useful information from the remaining point cloud. To address this issue, in addition to component masking, 
we also apply random masking on each remaining component separately. 
This can guide the model to gradually improve its understanding of the point cloud by reconstructing different components individually, 
ultimately achieving satisfactory performance.

\subsection{Component Semantic-enhanced Prompt-tuning}
We first introduce the classical classification head without proposed prompt-tuning.
During classification downstream tasks, the entire 3D architecture, except for the decoder, are fine-tuned.
Additionally, a randomly initialized class token $t_{cls}$ is concatenated with the tokens $T$ generated by the token embedding module and fed into the encoder for updating.
With the encoded $\{ t_{cls}^e,  T^e \}$, max pooling is applied to $T^e$ and acquire features $F_g \in \mathcal{R}^{1 \times C}$. 
The classification feature is $\{ t_{cls}^e, F_g \}$, and we use a non-linear layer to generate the final classification scores.

As shown in Figure~\ref{fig:4}(b), on the top of $\hat{p}$, we modifiy the feature before encoder by concatenating $\hat{p}$ with $\{t_{cls}$, $T\}$.
Then, after encoding, max pooling is applied to $\hat{p}^e$ and acquire semantic-enhanced features $\hat{p}^e_g \in \mathcal{R}^{1 \times C}$.
The new classifiction features is formulated as $\{ t_{cls}^e, \hat{p}^e_g, F_g \}$.

% %%%%%%%%%%%% Experiments

\section{Experiments}
In this section, we first present the pre-training setup and implementation details.
Then, to demonstrate the effectiveness of our modules, we evaluate the proposed modules using two baseline methods with four downstream tasks,
including synthetic object classification, real-world object classification, part segmentation and few-shot learning.
We also carry on ablation studies for our proposed approaches.
\subsection{Pre-training Setup}
We adopt the well-known ShapeNet~\cite{chang2015shapenet} for self-supervised point cloud pre-train,
which contains 57,448 synthetic point clouds with 55 object categories.

We adopt our modules on two baseline models, the transformer-based PointMAE~\cite{2022Masked} and the mamba-based PointMamba~\cite{liang2024pointmamba},
and we follow the same MAE architecture as these two methods for fair comparison.
For PointMAE, we adopt a typical
input resolution with 1024 points and divide inputs into n = 64 point patches. For the KNN
algorithm, we set k = 32. In the backbone, the encoder has 12 Transformer blocks while
the decoder has 4 Transformer blocks. Each Transformer block has 384 hidden
dimensions and 6 heads.
For PointMamba, the network architecture remains consistent with PointMAE,
but all the Transformers have been replaced with Mamba~\cite{gu2023mamba} structures.
More details are presented in supplementary materials.

\subsection{Effectiveness of Modules on Downstream Tasks}
To demonstrate the effectiveness of our approaches, we evaluate the proposed modules on two baseline methods with four downstream tasks.
In the Table~\ref{table:ScanObjectNN}, CSeM refer to our module for generating semantic-correlated masks and CSeP represents the proposed prompt-tuning strategy.

\begin{table*}[!t]
    % \vspace{-0.5em}
    \begin{center}
        % \setlength\tabcolsep{7pt}
        % \label{tab:1}
        \small
        \begin{tabular}{lccccc}
            \toprule
            % \hline
            \multirow{2}*{Method}                                & \multicolumn{3}{c}{ScanObjectNN}     & \multicolumn{2}{c}{ModelNet40}                                                                                                                            \\
            \cmidrule(lr){2-4} \cmidrule(lr){5-6}
            ~                                                    & $\mbox{OBJ-BG}$                       & $\mbox{OBJ-ONLY}$                     & $\mbox{PB-T50-RS}$                    & $\mbox{w/o Voting}$                 & $\mbox{w/ Voting}$                  \\
            \midrule
            % \cline{1-11}

            Point-BERT~\cite{yu2022point}                        & 87.43                                          & 88.12                                 & 83.07                                 & 92.7                                & 93.2                                \\
            % \cline{1-11}
            MaskPoint~\cite{liu2022masked}                       & 89.70                                          & 89.30                                 & 84.60                                 & -                                   & 93.8                                \\
            PointMAE~\cite{2022Masked}                           & 90.02                                          & 88.29                                 & 85.18                                 & 93.0                                & 93.8                                \\
            PointM2AE~\cite{zhang2022point}                      & 91.22                                          & 88.81                                 & \underline{86.43}                              & 93.4                       & 94.0                                \\
            ACT$^{\dag}$~\cite{dong2022autoencoders}                      & {93.29}                               & {91.91}                               & {88.21}                               & {93.7}                              & {94.0}                              \\
            ReCon$^{\dag}$~\cite{qi2023contrast}                          & {94.15}                               & {93.12}                               & {89.73}                               & {94.5}                              & {94.7}                              \\
            PointMamba~\cite{liang2024pointmamba}                & {90.71}                                        & {88.47}                               & {84.87}                               & {92.9}                              & {93.6}                              \\

            \midrule
            PointMAE~\cite{2022Masked} (baseline)                & 90.02                                          & 88.29                                 & 85.18                                 & 93.0                                & 93.8                                \\
            + CSeM (Ours)                                        & 90.90\textbf{\tiny\color{red}(+0.88)}          & 89.51\textbf{\tiny\color{red}(+1.22)} & 86.12\textbf{\tiny\color{red}(+0.94)} & 93.5\textbf{\tiny\color{red}(+0.5)} & 94.3\textbf{\tiny\color{red}(+0.5)} \\
            + CSeP (Ours)                                        & \underline{91.32}\textbf{\tiny\color{red}(+1.30)} & \underline{90.30}\textbf{\tiny\color{red}(+2.01)} & \textbf{86.64}\textbf{\tiny\color{red}(+1.46)} & \textbf{93.8}\textbf{\tiny\color{red}(+0.8)} & \textbf{94.5}\textbf{\tiny\color{red}(+0.7)} \\

            \midrule
            PointMamba~\cite{liang2024pointmamba}     (baseline) & {90.71}                               & {88.47}                               & {84.87}                               & {92.9}                              & {93.6}                              \\
            + CSeM (Ours)                                        & 91.63\textbf{\tiny\color{red}(+0.92)} & 89.55\textbf{\tiny\color{red}(+1.08)} & 85.61\textbf{\tiny\color{red}(+0.74)} & 93.3\textbf{\tiny\color{red}(+0.4)} & 94.0\textbf{\tiny\color{red}(+0.4)} \\
            + CSeP (Ours)                                        & \textbf{91.95}\textbf{\tiny\color{red}(+1.24)} & \textbf{90.59}\textbf{\tiny\color{red}(+2.12)} & 86.12\textbf{\tiny\color{red}(+1.25)} & \underline{93.6}\textbf{\tiny\color{red}(+0.7)} & \underline{94.2}\textbf{\tiny\color{red}(+0.6)} \\
            % \midrule
            % \multicolumn{4}{l}{Methods using cross-modal information or extra training data}                                         \\

            \bottomrule
        \end{tabular}
        % \vspace{-0.8em}
        % \vspace{-2mm}
        \caption{\textbf{Shape Classification on ScanObjectNN and ModelNet40 Datasets.} For ScanObjectNN dataset, we report the classification accuracy(\%) over the three subsets: OBJ-BG, OBJ-ONLY and the most challenging variant PB-T50-RS. For ModelNet40 dataset, we report the finetuning results with and without the voting trick.
        Methods with $^{\dag}$ are cross-modal methods. PointM2AE utilizes hierarchical encoder based on PointMAE.}
        \label{table:ScanObjectNN}
        % \vspace{-1em}
    \end{center}
    % \vspace{-6mm}
\end{table*}

\textbf{Shape Classification on a Real-world Dataset.}
To validate the effectiveness of the proposed model, we conduct experiments on ScanObjectNN~\cite{uy2019revisiting}dataset,
which consists of about 15,000 objects from 15 categories.
We evaluate our approaches effectiveness on PointMAE and PointMamba, and as shown in Table~\ref{table:ScanObjectNN}, we report
the overall accuracy under three experimental settings, OBJ-BG, OBJ-ONLY and PB-T50-RS.
All methods utilize the default data argumentation as the baseline.
When using the same pre-training dataset ShapeNet~\cite{chang2015shapenet},
our modules effectively improve the performance of the baseline models on this task.
Our modules improve the two baseline models by $2.01\%$ and $2.12\%$ respectively on the OBJ-ONLY split.
This demonstrates the effectiveness of our approach in capturing object semantics.
Additionally, on the split containing background OBJ-BG and the most challenge split PB-T50-RS,
our modules effectively improve the baseline performance in Real-world Dataset.

\textbf{Shape Classification on a Synthetic Dataset.}
In addition to the experiments conducted on a real-world dataset, we
perform experiments on a synthetic dataset, ModelNet40~\cite{wu20153d},
which consists of 12,311 clean 3D CAD models, covering 40 object categories.
For testing the fine-tuned model, we provide results with and without the
voting trick~\cite{liu2019relation}. The voting trick involves sampling multiple
point clouds for the same sample and making model predictions multiple times,
then aggregating the predictions through voting to obtain the final classification result.
As Shown in Table~\ref{table:ScanObjectNN}, our modules demonstrate effectiveness under both settings.
The results on the synthetic dataset further illustrate
that the performance of the pretrain model can be improved by masking complete components.

\begin{figure}[h] % h 选项表示将图片放在当前位置，而不是浮动到页面的顶部或底部
    \centering % 居中显示图片
    \includegraphics[width=0.48\textwidth]{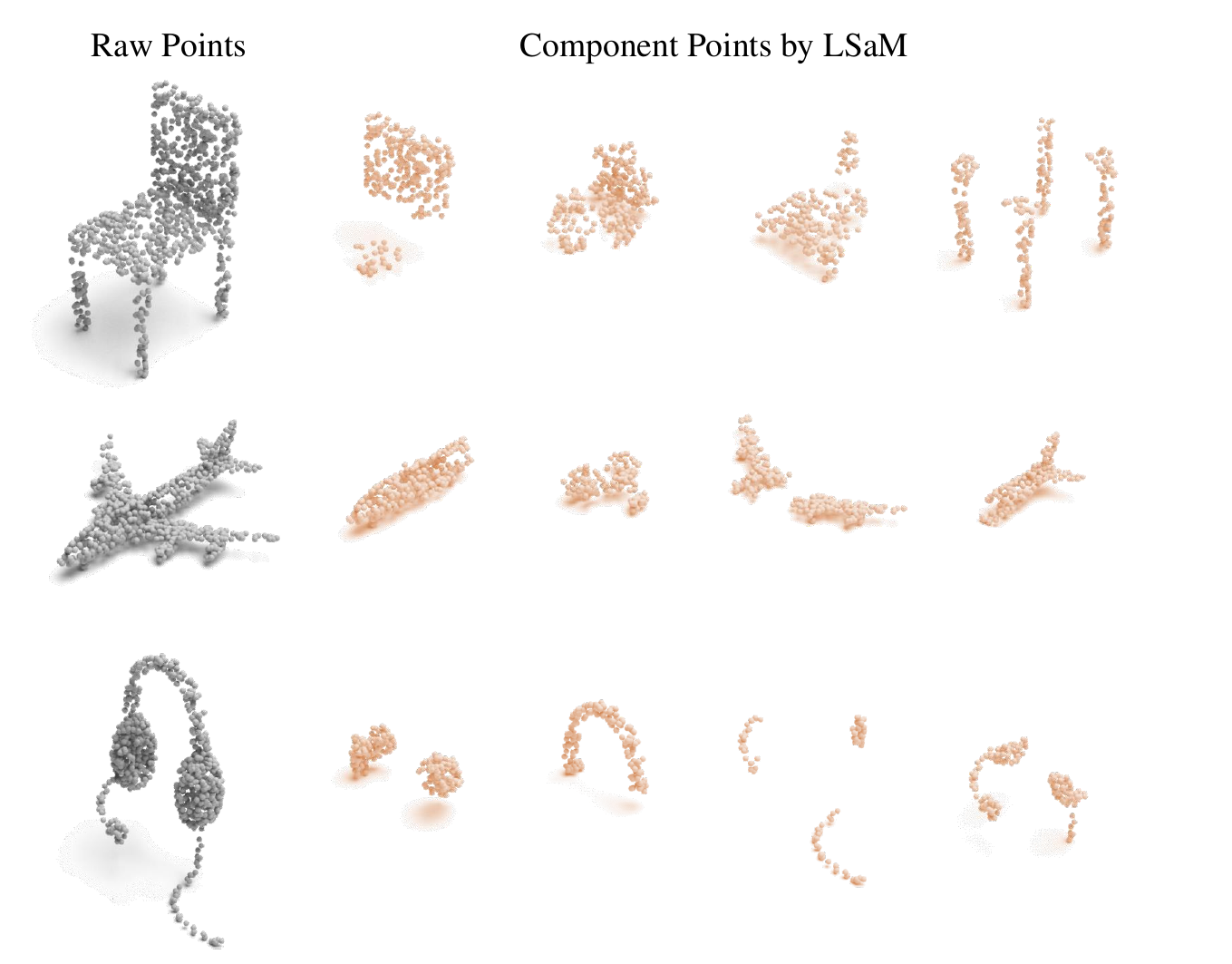} % 插入图片，并设置宽度为页面宽度的一半
    \vspace{-6mm}
    \caption{\textbf{Visualization of Components.} We visualized the components obtained through semantic grouping, which clearly show strong semantic relevance.
    }  % 添加图片标题
    \label{fig:5} % 添加图片标签，方便交叉引用
    % \vspace{-3mm}
\end{figure}

\begin{table}[!t]
    % \vspace{-0.5em}
    \begin{center}
        \setlength\tabcolsep{4pt}
        \small
        \label{table:ShapeNetPart}
        % \vspace{-0.8em}
        \begin{tabular}{lcc}
            \toprule
            % \hline
            {Method}                                             & Inst. mIoU                          & Cls. mIoU  \\
            \midrule
            % \cline{1-11}

            Point-BERT~\cite{yu2022point}                        & 85.6                                & 84.1            \\
            % \cline{1-11}
            MaskPoint~\cite{liu2022masked}                       & 86.0                                &  84.4             \\
            PointMAE~\cite{2022Masked}                           & 86.1                                &  84.1             \\
            ACT~\cite{dong2022autoencoders}                      & 86.1                                &  \underline{84.7}               \\
            PointMamba~\cite{liang2024pointmamba}                & 86.0                                &  84.4                 \\

            \midrule
            PointMAE~\cite{2022Masked}                           & 86.1                                      &  84.1           \\
            \textbf{+ Ours}                                        & \textbf{86.6}\textbf{\tiny\color{red}(+0.5)}       &  84.5\textbf{\tiny\color{red}(+0.4)}            \\

            \midrule
            PointMamba~\cite{liang2024pointmamba}                 & 86.0                                       &  84.4             \\
            \textbf{+ Ours}                                        & \underline{86.4}\textbf{\tiny\color{red}(+0.4)}        &  \textbf{84.9} \textbf{\tiny\color{red}(+0.5)}          \\
            % \midrule
            % \multicolumn{4}{l}{Methods using cross-modal information or extra training data}                                         \\
            \bottomrule
        \end{tabular}
        % \vspace{-2mm}
         \caption{\textbf{Part Segmentation on ShapeNetPart Dataset.} We report the mean IoU across all instances (Inst. mIoU) and the mIoU across all part categories (Cls. mIoU).}
        % \vspace{-1em}
    \end{center}
    % \vspace{-3mm}
\end{table}

\begin{table}[!t]
  % \vspace{-0.5em}
  \begin{center}
    \setlength\tabcolsep{3pt}
    \small
    % \vspace{-0.8em}
    % \vspace{-2mm}
    \begin{tabular}{lcccc}
      \toprule
      % \hline
      \multirow{2}*{Model} & \multicolumn{2}{c}{5-way}             & \multicolumn{2}{c}{10-way}                                                                                      \\
      \cmidrule(lr){2-3} \cmidrule(lr){4-5}
      ~                    & $\mbox{10-shot}$                       & $\mbox{20-shot}$                    & $\mbox{10-shot}$                    & $\mbox{20-shot}$                    \\
      \midrule
      % \cline{1-11}
      Point-BERT           & 94.6{\scriptsize $\pm$3.6}             & 93.9{\scriptsize $\pm$3.1}          & 86.4{\scriptsize $\pm$5.4}          & 91.3{\scriptsize $\pm$4.6}          \\
      % \cline{1-11}
      MaskPoint            & 95.0{\scriptsize $\pm$3.7}             & 97.2{\scriptsize $\pm$1.7}          & 91.4{\scriptsize $\pm$4.0}          & 92.7{\scriptsize $\pm$5.1}          \\
      MAE3D                & 95.2{\scriptsize $\pm$3.1}             & 97.9{\scriptsize $\pm$1.6}          & 91.1{\scriptsize $\pm$4.6}          & {95.3{\scriptsize $\pm$3.1}}        \\
      PointMAE             & 96.3{\scriptsize $\pm$2.5}             & 97.8{\scriptsize $\pm$1.8}          & 92.6{\scriptsize $\pm$4.1}          & 93.4{\scriptsize $\pm$3.5}          \\
      PointM2AE            & 96.8{\scriptsize $\pm$1.8}             & 98.3{\scriptsize $\pm$1.4}          & 92.3{\scriptsize $\pm$4.5}          & 95.0{\scriptsize $\pm$3.0}          \\
      ACT                  & {96.8{\scriptsize $\pm$2.3}}          & {98.0{\scriptsize $\pm$1.4}}       & {93.3{\scriptsize $\pm$4.0}}       & {95.6{\scriptsize $\pm$2.8}}       \\
      ReCon                & {97.3{\scriptsize $\pm$1.9}}          & {98.0{\scriptsize $\pm$1.4}}       & {93.3{\scriptsize $\pm$3.9}}       & {95.8{\scriptsize $\pm$3.0}}       \\
      PointMamba      & {95.0{\scriptsize $\pm$2.3}}          & {97.3{\scriptsize $\pm$1.8}}       & {91.4{\scriptsize $\pm$4.4}}       & {92.8{\scriptsize $\pm$4.0}}       \\ 
      % \textbf{Ours}                       & \underline{97.0{\scriptsize $\pm$2.4}}    & 98.5{\scriptsize $\pm$1.3}                & \underline{93.9{\scriptsize $\pm$3.9}}    & \underline{95.6{\scriptsize $\pm$3.8}}    \\
      % \hline
      % \multicolumn{5}{l}{Methods using cross-modal information or extra training data}      
      \midrule
      % CorssNet~\cite{}                    & {\color{gray} 96.8{\scriptsize $\pm$2.3}} & {\color{gray} 98.0{\scriptsize $\pm$1.4}} & {\color{gray} 93.3{\scriptsize $\pm$4.0}} & {\color{gray} 95.6{\scriptsize $\pm$2.8}} 
      PointMAE    (baseline)         & 96.3{\scriptsize $\pm$2.5}             & 97.8{\scriptsize $\pm$1.8}          & 92.6{\scriptsize $\pm$4.1}          & 93.4{\scriptsize $\pm$3.5}          \\

      \textbf{+ Ours}        & \textbf{96.8{\scriptsize $\pm$2.2}} & \textbf{98.0{\scriptsize $\pm$1.5}} & \textbf{93.0{\scriptsize $\pm$4.5}} & \textbf{95.3{\scriptsize $\pm$2.7}} \\

      \midrule
      PointMamba      & {95.0{\scriptsize $\pm$2.3}}          & {97.3{\scriptsize $\pm$1.8}}       & {91.4{\scriptsize $\pm$4.4}}       & {92.8{\scriptsize $\pm$4.0}}       \\ 

      \textbf{+ Ours}        & \textbf{96.0{\scriptsize $\pm$2.0}} & \textbf{98.3{\scriptsize $\pm$1.7}} & \textbf{92.0{\scriptsize $\pm$4.5}} & \textbf{93.6{\scriptsize $\pm$3.8}} \\
      % \\

      \bottomrule
    \end{tabular}
    \caption{\textbf{Few-shot Classification on ModelNet40.} We report mean and standard deviation over ten runs.}
    % \vspace{-1em}
    \label{table:fewshot}
  \end{center}
  % \vspace{-5mm}
  
\end{table}

\begin{table}[ht]
    \begin{center}
        \small
        \setlength\tabcolsep{2pt}
        % \small
        % \vspace{1em}
        \begin{tabular}{ccccc}
            \toprule
            \multirow{2}{*}{CrossAttn} & \multirow{2}{*}{ContrastL } & \multirow{2}{*}{ProtoRecon}  & \multicolumn{2}{c}{Overall Accuracy}      \\
            \cmidrule{4-5}                     &                               &                                            & OBJ-ONLY                    & ModelNet40   \\
            \midrule
            \ding{55}                          & \ding{55}                     & \ding{55}                                      & 88.76                           & 92.95 \\
            \ding{51}                          & \ding{55}                     & \ding{55}                                       & 88.55                          & 93.08 \\
            \ding{51}                          & \ding{51}                     & \ding{55}                                       & 87.89                          & 92.69  \\
            \ding{51}                          & \ding{51}                     & \ding{51}                                       & \textbf{89.51}                           & \textbf{93.51}          \\
            \bottomrule
        \end{tabular}
    \caption{\textbf{Evaluation of PCSM.} Short \textbf{CrossAttn} represents Cross Attention module,
    \textbf{ContrastL} represents perfom contrastive learning between prototypes, \textbf{ProtoRecon} represents reconstruct point clouds base on prototypes.}
     \label{table:pcsm}
    \end{center}
    % \vspace{-3mm}
\end{table}

\begin{table}[ht]
    \begin{center}
        \small
        \setlength\tabcolsep{2pt}
        % \small
        % \vspace{-2mm}
        \begin{tabular}{ccccc}
            \toprule
            \multirow{2}{*}{RandM} & \multirow{2}{*}{RandBM } & \multirow{2}{*}{CSeM} &  \multicolumn{2}{c}{Overall Accuracy}      \\
            \cmidrule{4-5}                     &                               &                                             & OBJ-ONLY                    & ModelNet40   \\
            \midrule
            \ding{55}                          & \ding{51}                     & \ding{55}                                      & 88.28                            & 92.86 \\
            \ding{51}                          & \ding{55}                     & \ding{55}                                        & 88.79                          & 93.09  \\
            \ding{55}                          & \ding{55}                     & \ding{51}                                      & \textbf{89.51}                           & \textbf{93.51}          \\
            \bottomrule
        \end{tabular}
        \caption{\textbf{Evaluation of different masking strategy.} Short \textbf{RandM} represents global random mask,
        \textbf{RandBM} represents random block mask, \textbf{CSeM} represents semantic-correlated random mask.}
        % \vspace{1em}
        \label{table:mask}
    \end{center}
    % \vspace{-5mm}
\end{table}

\begin{table}[ht]
    \begin{center}
        \small
        \setlength\tabcolsep{3pt}
        % \small
       
        \begin{tabular}{ccccc}
            \toprule
            \multirow{2}{*}{Prompt} & \multirow{2}{*}{MaxP} & \multirow{2}{*}{Concat}  & \multicolumn{2}{c}{Overall Accuracy}      \\
            \cmidrule{4-5}                     &                               &                                               & OBJ-ONLY                    & ModelNet40   \\
            \midrule
            \ding{55}                          & \ding{55}                     & \ding{55}                                     & 88.98                            & 93.08 \\
            \ding{51}                          & \ding{55}                     & \ding{55}                                      & 89.35                           & 93.24 \\
            \ding{51}                          & \ding{55}                     & \ding{51}                                      & 89.93                         & 93.66  \\
            \ding{51}                          & \ding{51}                     & \ding{51}                                     & \textbf{90.30}                 & \textbf{93.78}  \\
            \bottomrule
        \end{tabular}
        \caption{\textbf{Evaluation of CSeP.} Short \textbf{Prompt} represents that adopt prototypes as prompts,
        \textbf{MaxP} represents that perform maxpooling before concatenate prototypes with classification head's input, \textbf{Concat} represents concatenate prototypes with head's input.}
        % \vspace{1em}
        \label{table:CSeP}
        % \vspace{-2mm}
    \end{center}
    % \vspace{-3mm}
\end{table}

% \begin{table}[ht]
%     \begin{center}
%         \small
%         \setlength\tabcolsep{5pt}
%         % \small
%         \begin{tabular}{cccc}
%             \toprule
%             \multirow{2}{*}{ABAB} & \multirow{2}{*}{AABB } & \multicolumn{2}{c}{Overall Accuracy}      \\
%             \cmidrule{3-4}                     &                           & OBJ-ONLY                    & ModelNet40   \\
%             \midrule
%             \ding{55}                          & \ding{51}                & 88.98                           & 92.73 \\
%             \ding{51}                          & \ding{55}                & \textbf{89.51}                          & \textbf{93.51} \\
    
%             \bottomrule
%         \end{tabular}
%          \caption{\textbf{Evaluation of Cross-attention Module in PCSM.} \textbf{ABAB} first uses features to update prototypes, then uses prototypes to update features, both via cross-attention. \textbf{AABB} updates prototypes twice with features, then updates features twice with prototypes.}
%         % \vspace{1em}
%         \label{table:cross}
%         % \vspace{-2mm}
%     \end{center}
%     % \vspace{-5mm}
% \end{table}

\textbf{Part Segmentation on ShapeNetPart Dataset.}
We conduct part segmentation experiments on the challenging
ShapeNetPart~\cite{yi2016scalable} dataset,
which comprises 16880 models with 16 different shape categories and 50 part labels.
The proposed paradigm boost the performence of 3D pre-trained on the dataset, which is one of the hardest task.
The "PointMAE + CLIP" combination outperforms all methods in various experimental settings, 
demonstrating that vision semantic prompts can effectively introducing part semantics.

\textbf{Few-shot Classification.}
To evaluate the effectiveness of the proposed modules with limited finetuning data, we conduct
experiments for few-shot classification on ModelNet40. As
shown in Table~\ref{table:fewshot}, the proposed modules improve the performance of baselines
for all four settings. The results illustrate that
our approach can augment pretrain models generalization capabilities by enhancing local semantics.

\subsection{Ablation studies}
To investigate the effectiveness of each component and design, we conduct ablation studies on
ScanObjectNN with the OBJ-ONLY variant and ModelNet40 without voting. We apply PointMAE as the baseline.

\textbf{Effect of PCSM.}
To illustrate the effect of modules in PCSM for capturing reasonable local semantics,
we conduct ablation experiments on the PCSM.
As shown in Table~\ref{table:pcsm}, it can be observed that the masks generated by random prototypes unable to improve the network's performance.
Without additional constraints, this approach causes the attention regions of the prototypes to be randomly distributed.
The introduction of contrastive learning leads to a performance drop.
Without guidance, even though prototypes can focus on different local blocks, but fail to model proper local semantics.
To guide the prototypes in modeling reasonable local semantics, we employ prototype-based point cloud reconstruction.
Additionally, we incorporate position embedding to further enhance the prototypes ability to capture local semantic details, which boosts the performance of models.

\textbf{Effect of Different Masking Strategies.}
To assess the impact of different masking strategies,
we applied three strategies with the same backbone, the origin masking strategy (Global Random Mask)
randomly mask local structures (Random Block Mask) and random mask based on local semantics (Ours).
As shown in Table~\ref{table:mask}, without local prototypes guidance,
random block mask in the point cloud is ineffective,
as it fails to address the issue of residual local structures.
Meanwhile, unlike the global random mask which overlooks differences between various components of the point cloud,
our masking strategy improves the performance by enabling the model to extract local-representative features.
This enhances the pre-trained model's ability to capture local semantics and improves models performance.
As shown in Figure~\ref{fig:5}, we present the visualization of components segmented by pre-trained model,
our approach can obtain point cloud components that show strong semantic relevance.

\textbf{Effect of CSeP.}
We conduct ablation experiments on the CSeP.
As shown in Table~\ref{table:CSeP}, 
we first use randomly initialized prototypes during finetuning without pre-training,
which only adds random noise to the input without providing meaningful information.
After employing pre-trained prototypes, which introduces useful local semantic information,
the performance saw only a marginal improvement.
after prototypes are updated by the encoder in finetune,
we concatenate them into the input of the classification head after max-pooling

% \textbf{Effect of Cross-attention Structures in PCSM.}
% As our core module, we conduct an exploration of the specific architecture of the cross-attention module in PCSM.
% As shown in the Table~\ref{table:cross}, the ABAB architecture 
% We analyze the ABAB structure through alternating updates,
% can simultaneously optimize the prototypes and enhance the semantic representation of features,
% leading to better prototype optimization.

% %%%%%%%%%%% Conclusion
\section{Conclusion}
In this paper, 
% we identify a key issue in MAE-based point cloud understanding methods :
% the methods fail to capture complete component semantics during pre-training.
we propose the PCSM and the CSeM, which enable model to mask complete point cloud components,
effectively improving the effectiveness of pre-trained models in downstream tasks.
Additionally, on the top of local semantic-enhanced portotypes, 
we adopt these prototypes as prompts to improve the performance in downstream tasks.
Extensive experiments demonstrate the effectiveness of our approach.

\bibliographystyle{named}
\bibliography{ijcai25}

\end{document}